# REPAIRING PEOPLE TRAJECTORIES BASED ON POINT CLUSTERING

Duc-Phu CHAU[(1)(2)], François BREMOND[(1)], Etienne CORVEE[(1)] and Monique THONNAT[(1)]

[(1)] *Pulsar, INRIA*
*2004 route des Lucioles, BP93, 06902 Sophia Antipolis Cedex, France*
*{Duc-Phu.Chau, Francois.Bremond, Etienne.Corvee, Monique.Thonnat}@sophia.inria.fr*
*http://www-sop.inria.fr/pulsar*
[(2)] *Department of Technology, Phu Xuan Private University*
*176 Tran Phu, Hue, Thua Thien Hue, Vietnam*
*http://www.phuxuanuni.edu.vn*

Abstract:   This paper presents a method for improving any object tracking algorithm based on machine learning. During the training phase, important trajectory features are extracted which are then used to calculate a confidence value of trajectory. The positions at which objects are usually lost and found are clustered in order to construct the set of 'lost zones' and 'found zones' in the scene. Using these zones, we construct a triplet set of zones i.e. 3 zones: In/Out zone (zone where an object can enter or exit the scene), 'lost zone' and 'found zone'. Thanks to these triplets, during the testing phase, we can repair the erroneous trajectories according to which triplet they are most likely to belong to. The advantage of our approach over the existing state of the art approaches is that (i) this method does not depend on a predefined contextual scene, (ii) we exploit the semantic of the scene and (iii) we have proposed a method to filter out noisy trajectories based on their confidence value.

## 1   INTRODUCTION

Nowadays video surveillance systems are applied worldwide in many different sites such as parking lots, museums, hospitals and railway stations. Video surveillance helps a supervisor to overlook many different camera fields of views from the same room and to quickly focus on abnormal events taking place in the control space. However one question arises: how a security officer can analyze in real time and simultaneously dozens of monitors with a minimum rate of missing abnormal events? Moreover, the observation of screens for a long period of time becomes boring and draws the supervisor's attention away from the events of interest. The solution to this issue lies in three words: intelligent video monitoring.

Intelligent video systems belong to the domain that studies the content of a video. This term expresses a research direction fairly large, and is applied in different fields: for example in robotics and homecare. In particular, a lot of research and work in this area are already achieved in video surveillance. This paper focuses in one of the most important issue of intelligent video monitoring: mobile object tracking.

There are many methods proposed to track mobile objects [1,2,3,6,7,8]. In [7] the authors propose to use Kalman Filter combined with optimization techniques for data association in order to filter and robustly manage occlusions and non-linear movements. In [1], a method for tracking multiple moving objects is presented, using particle filters to estimate the object states based on joint probabilistic data association filters. In [6], the authors also propose to use particle filters that are quite flexible as they can approximate any probability distribution with a large set of particles, and allow non-linear dynamics to be encapsulated. All these works listed above have obtained satisfactory results, but complex and long situations of occlusion are not addressed.

To solve the occlusion problem, some researches [4,5] have focused on modelling the scene in order to improve the tracking algorithm. The main idea consists in providing the interesting information in the scene such as: the positions, directions of paths (i.e tracked objects), the sensitive zones in the scene where the system can lose object tracks with a high probability, the zones where mobile objects appear and disappear usually… These elements can help the system to give better prediction and decision on object trajectory. There are two possible ways to model a scene either using machine learning techniques or by hand. With machine learning, the modelling cost is low, but the modelling algorithm has to insure the quality and the precision of the constructed scene model. For instance, the authors in [5] have presented a method to model the paths in the scene based on the detected trajectories. The construction of paths is performed automatically using an unsupervised learning technique based on trajectory clustering. However, this method can only be applied to simple scenes where only clear routes are defined. The criteria for evaluating a noisy trajectory are mostly based on trajectory duration. Fernyhough et al [4] use the same model for learning automatically object paths by accumulating the trace of tracked objects. However, it requires full trajectories, it cannot handle occlusions and the results depend on the shape and size of the objects, as they are detected on the 2D image plane.

To solve these problems, we use machine learning in order to extract automatically the semantic of the scene. We also propose a method to calculate the confidence value of trajectories. This value is used to filter the noisy trajectories before the learning process, and also to learn some special zones (eg. entrance and exit zones) in the scene with which the system can recover a trajectory after losing it.

The rest of the paper is organized as follows. In the next section, a description of the approach working steps i.e. the machine learning stage and the testing phase is given. The experimentation and validation of the approach are presented in section 3. A conclusion is given in the last section as well as some propositions to improve our algorithm for better trajectory repairing.

## 2  OVERVIEW OF THE APPROACH

### 2.1  Features for Trajectory Confidence Computation

The proposed approach takes as input the track objects obtained by any tracking algorithm. To validate the proposed algorithm, we have used a region based tracking algorithm [anonymous] where moving regions are detected by reference image subtraction. One of the most important problems in machine learning is to determine the suitable features for describing the characteristics of a trajectory.

We aim at extracting features that enable the distinction between noisy trajectories and true trajectories of real mobile objects. In this paper, we propose and define 9 features:

1. An entry zone feature is activated when an object enters the scene in the entry zone e.g. the zone around a door.

2. An exit zone feature is activated when an object disappears in an exit zone. It is a zone from where the object can leave the scene.

3. Time: the lifetime of the trajectory.

4. Length: the spatial length of the trajectory.

5. Number of times the mobile object is classified as a 'person'. An object is classified according to its 3D dimension and a predefined 3D object model such as a person. This number is directly proportional to its trajectory's confidence value.

6. Number of times that the trajectory is lost.

7. Number of neighbouring mobile objects at four special temporal instants. Here we count the number of mobile objects near the considered mobile object when it has been (1) detected for the first time, (2) lost, (3) found (if previously lost) and (4) when the trajectory ends. This feature is used to evaluate the potential error when detecting an object. The greater this number of neighbours is, the lower the confidence of the trajectory.

8. Number of times the mobile object changes its size according to a predefined dimension variation threshold. The too large variation of a mobile object's size will penalize objects in having a high confidence trajectory.

9. Number of times the mobile object changes spatial direction. The usual behaviour of people in subway stations is to go in straight direction from one location to another e.g. from the gates to the platform. When this feature is high, the trajectory confidence is low.

In total, nine features defined above are used to characterize the confidence of a detected trajectory. For calculating this confidence value a normalisation phase is necessary. The values of features 5 and 8 are normalised by the time length of the corresponding trajectory.

The features 1 and 2 are Booleans: 0 (non-activated) or 1 (activated). The value of the other features (3, 4, 6, 7 and 9) are normalised as follows:

$$f_i = \frac{FV_i - \mu_i}{\sigma_i} \qquad (1)$$

where,
$FV_i$: the value of feature $i$ where $i = \{3,4,6,7,9\}$
$\mu_i$: the average of feature $i$ value for all trajectories processed in the learning stage
$\sigma_i$: the variance of feature $i$ value for all trajectories processed in the learning stage
$f_i$: the new value of feature $i$ after normalisation

The confidence value of a trajectory is calculated by the following formula:

$$CV = \sum_{i=1}^{5}(w_i * f_i) + \sum_{i=6}^{9}(w_i *(1 - f_i)) \qquad (2)$$

where,
$CV$ = confidence value of trajectory considered
$w_i$ = the weight (importance) of feature $i$
$f_i$ = the value of feature i after normalisation

The first 5 features are the ones being directly proportional to the confidence value; the last features are the ones being inversely proportional to the confidence value.

## 2.2 Learning Feature Weights with Genetic Algorithm

We learn the nine weights associated to the trajectory features with a genetic algorithm. Firstly, we select the first 300 trajectories in a video for training set and also the trajectory feature values. In order to find the importance (weight) of features, we associate a ground truth of each trajectory. The ground truth is manually defined as the global confidence of a trajectory and these confidence values are in the interval [0..1]. We have defined four levels in this interval to classify the trajectories in 4 classes:
− complete trajectory: ground truth >= 0.8
− incomplete trajectory: does not start or does not end in an In/Out zone, 0.5 <= ground truth < 0.8
− unreliable trajectory: does not start and does not end in an In/Out zone, 0.2 <= ground truth < 0.5
− noise: does not correspond to the trajectory of a person, ground truth < 0.2

The problem consists in learning the feature weights in order to optimize the correspondences between ground truth values and the confidence values calculated with the weights. There are many methods to learn the weights. Here we have selected a genetic algorithm because it economizes the time cost (we have up to 9 features), and it is effective.

### 2.2.1 Mutation and Cross-over for the Genetic Algorithm

We call an individual the set of 9 values representing the weights of trajectory features that need to be learnt.

The fitness of each individual with respect to the ground truth is used to calculate the difference (error) between the individual value and the ground truth value. In this case, the fitness value is calculated with the following formula:

$$\phi = \sum_{i=1}^{300} |GT(i) - CV(i)| \qquad (3)$$

Where,
$\phi$: fitness of the considered individual.
$GT(i)$: The ground truth value of trajectory $i$.
$CV(i)$: The confidence value of trajectory $i$ calculated by the considered weight set.

The better individual is one with the lower value of fitness.

In order to explore different distributions of individuals, we define two operators 'Mutation' and 'Cross-Over'. The mutation transformation processes each individual, with probability 30%. Once an individual is selected for the mutation, one feature weight position for this individual is randomly chosen. All the weight values of this individual from this position to the end will be changed by new random values.

The cross-over transformation is performed for each individual, with probability 80%. Two individuals are selected to change their weights by cross over. One feature weight position is randomly chosen for the two individuals. All weight values of these individuals from this position to the end will be swapped.

After performing mutation or cross-over operators, the sum of weights of each individual can be greater or less than 1. Each feature weight value will be divided by the sum of all values of that individual, which is necessary to ensure that this sum is always 1.

### 2.2.2 The genetic algorithm

We have created a first individual generation including 5000 individuals. For each generation, we perform the cross-over and mutation transformations in order to create a new generation. This production process is performed until the fitness of the best individual we obtain is less than a given threshold.

## 2.3 Types of zones used in a scene

In our approach, we want to exploit the semantic of the scene. To do this, we define several zone types.
- entry zone: zone where the object can enter the scene.
- exit zone: zone where the object can leave the scene.
- IO zone: zone where the object can enter and also leave the scene.
- lost zone: zone, which is not in the same time an exit or IO zone, where the tracking loses usually the object.
- found zone: zone, which is not in the same time an entry or IO zone, where the tracking detects usually new object.
- lost-found zone: is a zone having both characteristics "lost" and "found".

Fig 1 and Fig 2 show the description of the different zone types.

## 2.4 Zone Learned by Clustering

The entry, exit, IO zones are defined manually and this construction is based in our case on the subway zones. Inversely, the lost, found, lost-found zones are constructed automatically using machine learning. To learn these zones, the system runs the tracking process for getting trajectory data. The coordinates where the system loses or founds the tracked people, are marked in order to construct lost zones and found zones. A clustering process based on people 3D positions is realised. We chose Kmean algorithm for clustering and a first task is to determine the number of clusters which is done by expectation maximization (EM) algorithm.

The results of the clustering process for the lost zones are displayed in figure 3. In the figure 4, the outline of "lost zones" is drawn in yellow and the red zones are the entry, exit and IO zones.

```xml
<Zone ident = "9" name = "ZoneIOLeftTop" plane_name = "ground">
  <Properties_list>
      <Property name = "In_out_zone:Entry"/>
  </Properties_list>
  <Outline_list>
      <Point x="-830.0" y="-350.0" z = "0"/>
      <Point x="-300.0" y="-350.0" z = "0"/>
      <Point x="-300.0" y="-100.0" z = "0"/>
      <Point x="-830.0" y="-100.0" z = "0"/>
  </Outline_list>
</Zone>
```

Figure 1: Description of an entry zone

```xml
<Zone ident = "2" name = "ZoneLearning0" plane_name = "ground">
  <Properties_list>
      <Property name = "Lost_found_zone:Yes"/>
  </Properties_list>
  <Outline_list>
      <Point x="-2046.000000" y = "12.000000" z="0" />
      <Point x="-2046.000000" y = "778.000000" z="0" />
      <Point x="-1402.000000" y = "778.000000" z="0" />
      <Point x="-1402.000000" y = "12.000000" z="0" />
  </Outline_list>
</Zone>
```

Figure 2: Description of a lost-found zone

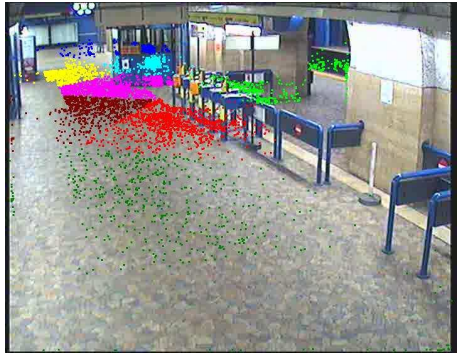 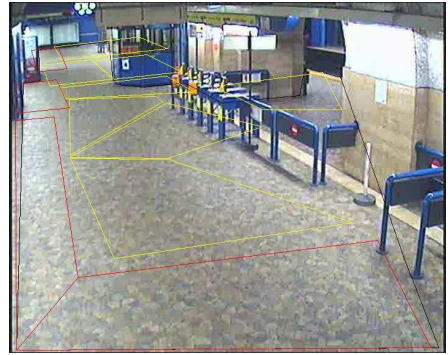

Fig 3. Clustering results for 8 lost zones. Each colour represents a cluster.

Fig 4. The yellow polygons show the outline of the learned lost zones. The red polygons show the outline of the entry zone, exit zone and IO zones.

### 2.5 Zone Triplet Calculation

A zone triplet is a structure for the system to repair lost trajectories. The system is using complete trajectories to repair similar but lost trajectories and thus to build the zone triplets. Once the zones are generated, the zone triplets can be constructed. A zone triplet is a set of 5 values (start Tzone, lost Tzone, found Tzone, minimum time, maximum time). "Start Tzone" is entry or IO zone where trajectories begin. "Lost Tzone" is the first lost zone or the first lost-found zone where the complete trajectories pass. "Found Tzone" is the first found zone or lost-found zone where the complete trajectories pass just after passing a lost Tzone. Found zone has to be searched only after the object has left its lost Tzone. Minimum time is calculated by the difference between the instant when entering the found Tzone and the instant when exiting the lost Tzone. Maximum time is calculated by the difference between the instant when leaving the found Tzone and the instant when entering the lost Tzone.

A complete trajectory is a trajectory whose confidence value is greater than a given threshold and starts from an entry or IO zone. Only complete trajectories passing successively through start, lost and found Tzones can be used for building a zone triplet. All zone triplets are ordered by the number of complete trajectories passing through them. A zone triplet with a greater number of trajectories will be ordered with higher priority. The minimum and maximum time are calculated as the average of the correspondent values for the associated complete trajectories. The next section will present how the triplets are used to repair the lost trajectories.

### 2.6 Repairing Lost Trajectories

Zone triplets are used to repair lost trajectories. When the system detects an object (considered as lost) that appears in an abnormal location (not in an IO zone or in an "entry zone"), we verify whether it appears in a found zone or in a lost-found zone ZF. If this is the case, we search among the triplets constructed in the previous step the triplet with the "found Tzone" ZF, such as <ZS, ZL, ZF, Mi, Ma>. If several triplets can be associated to the lost object, the system will chose the triplet having the highest priority. After that, the system searches for a lost trajectory that begins in ZS, that is lost in ZL and with a temporal interval between the lost instant up to current time which is greater than Mi and lower than Ma. Once a trajectory is found, the system fuses this trajectory with the trajectory of the lost object that has just been detected.

# 3 EXPERIMENTATION AND VALIDATION

Experiments have been performed with the videos of a European project (hidden for anonymous reason). This project aims at tracking people in subway stations and learning their behaviours. These videos are specially interesting due to the fact that people motions in the scene are diverse. There are not clear paths and the mobile objects can evolve anywhere. For that, the semantic exploitation of the scene is hard and required machine learning techniques. We have carried out the clustering algorithm on a video of 5 hours, and computed over 8000 trajectories. Just only 23 trajectories among them have been chosen to construct triplets, and 12 triplets have been found. The system has detected 340 lost trajectories that can be solved by the proposed repair algorithm. In order to evaluate the benefits of the algorithm, we have calculated the confidence value before and after fusing the trajectories. In the 340 detected cases, there were up to 337 cases where the confidence value increases. The system has wrongly repaired some lost trajectories when performing the fusion task. These errors have happened when several trajectories get lost at the same time but with different triplets. In this case, a lost object will be associated to the zone triplet with higher priority which can be an incorrect association.

The system has successfully filtered noisy trajectories thanks to their confidence value. The trajectories having a confidence value lower than a given threshold have been considered as noisy trajectories

The table 1 is a summary of data in two cases: apply and not apply the algorithm. We can see an increase of the complete trajectory number and a decrease of the incomplete trajectory and noise number. The system also detects successfully 4550 noisy trajectories in the first case and 4481 noisy trajectories in the second one. There is a decrease of 340 trajectories in total after applying the algorithm because there are 340 fusion cases.

In this section, we want to show some experimentation results. In figure 5, the image a) is captured before the system loses the trajectory with id 3 (cyan colour). The image b) is captured 2s later. The person is detected, but the system cannot recognize that this is the same person than the one in previous frames (with id 5, yellow colour). Whereas, in the image c) thanks to the proposed approach, the system is able to repair the lost trajectory, and the person is detected successfully up to the end.

We can see another example in figure 6. In this example, the system can track successfully the trajectory of person 2 (pink colour).

**Table 1.** Summary of results in two cases: with or without the proposed algorithm.

|  | Without the algorithm | | With the algorithm | |
| --- | --- | --- | --- | --- |
|  | Number | Percentage (%) | Number | Percentage (%) |
| Complete trajectories | 758 | 9.0 | 795 | 9.9 |
| Incomplete trajectories | 3086 | 36.8 | 2778 | 34.5 |
| Noise | 4550 | 54.2 | 4481 | 55.6 |
| Total | 8394 | 100 | 8054 | 100 |

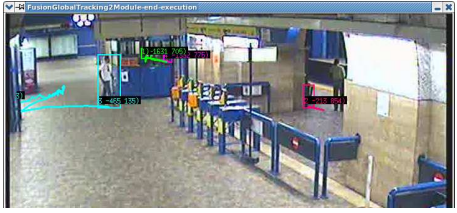

Fig 5a)

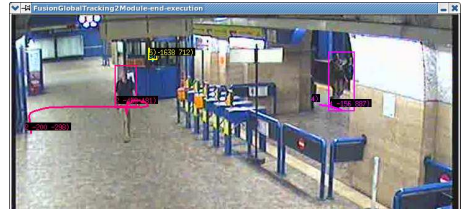

Fig 6a)

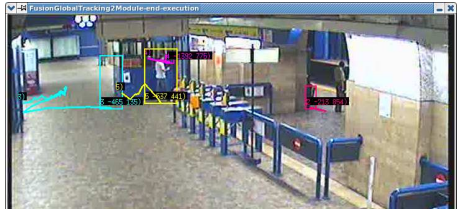

Fig 5b)

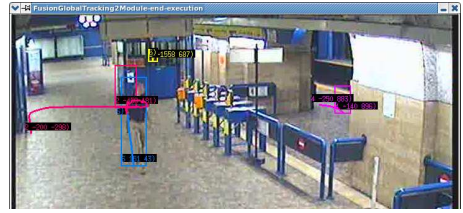

Fig 6b)

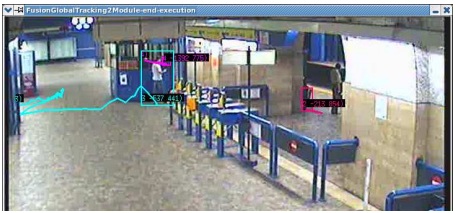

Fig 5c)

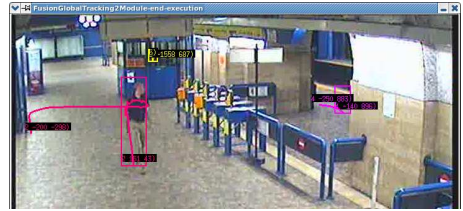

Fig 6c)

Fig 5. Images captured before and after trajectory fusion at t = 711
a) t = 709
b) t = 711 before trajectory fusion
c) t = 711 after trajectory fusion

Fig 6. Images captured before and after trajectory fusion at t = 903
a) t = 901
b) t = 903 before trajectory fusion
c) t = 903 after trajectory fusion

## 4 CONCLUSION

This paper presents a method to repair the trajectories of objects in videos. The principle of this approach is to use machine learning stage to exploit the semantic of the scene. The zones defined as lost zone, found zone and lost-found zone are constructed by the clusters of 3D object positions on the ground where the system can lose only, find only and lose or find respectively the trajectories. Nine trajectory features are extracted and used to calculate a confidence value of trajectory. The confidence value is used to filter out noisy trajectories. The best trajectories are used to construct zone triplets. A zone triplet is a 'representation' of a path describing the complete paths of people in a scene where trajectories can be lost. Zone triplets are suitable to the difficult conditions where there are not any clear paths such as in many areas of subway stations as shown in the experimentation section. Moreover, these triplets can detect the lost trajectories to be fused and repaired.

However, the approach still encounters problems that have to be considered in order to better repair trajectories. Although the use of zone triplets succeeded in repairing some trajectories, not all repaired trajectories were obtained. This is due to the complexity of the people activities in a scene and the large amount of short and incomplete trajectories to be fused. Although the work experiences some weak points, this method opens a new approach to improve any tracking algorithm and these initial results are promising for future studies. For instance, we can first increase the number of zones defined in triplet to four or five so that trajectory fusion becomes more versatile. Second, the first zone of the zone triplet does not have to be a start zone, but could be of any zone type. This start zone defined as a first zone was used to limit the space for searching trajectories to be fused and hence limits the repair process flexibility.